\documentclass[10pt, conference]{IEEEtran}
\IEEEoverridecommandlockouts

\usepackage{amsmath,amssymb,amsfonts}
\usepackage{graphicx}
\usepackage{hyperref}
\usepackage{caption, subcaption}
\usepackage[ruled,linesnumbered]{algorithm2e}
\usepackage{booktabs,multirow,threeparttable}

\newcommand*{\figureref}[1]{
Figure \ref{#1}
}
\newcommand*{\algorithmref}[1]{
Algorithm \ref{#1}
}

\newcommand*{\E}{\mathbb{E}}
\newcommand*{\Sc}{\mathcal{S}}
\newcommand*{\Ac}{\mathcal{A}}
\newcommand*{\nt}{\nabla_{\theta}}
\newcommand*{\pt}{\pi_\theta}
\newcommand*{\ptold}{\pi_{\theta_\mathrm{old}}}
\newcommand*{\tv}{{\theta^v}}
\newcommand*{\tvold}{{\theta^v_\mathrm{old}}}

\begin{document}

\title{Proximal Policy Optimization with Mixed Distributed Training}

\author{
	\IEEEauthorblockN{Zhenyu Zhang\IEEEauthorrefmark{1}, Xiangfeng Luo\IEEEauthorrefmark{1}\IEEEauthorrefmark{2}\thanks{\IEEEauthorrefmark{2}corresponding author}, Tong Liu\IEEEauthorrefmark{1}, Shaorong Xie\IEEEauthorrefmark{1}, Jianshu Wang\IEEEauthorrefmark{1}, Wei Wang\IEEEauthorrefmark{1}, Yang Li\IEEEauthorrefmark{1} and Yan Peng\IEEEauthorrefmark{1}}
	
	\IEEEauthorblockA{\IEEEauthorrefmark{2}Shanghai Institute for Advanced Communication and Data Science \\ \IEEEauthorrefmark{1}School of Computer Engineering and Science, Shanghai University, Shanghai, China}

	\IEEEauthorblockA{\{zhenyuzhang, luoxf,	tong\_liu, srxie, ces\_keavnn, familywei, abluceli, pengyan\}@shu.edu.cn}
}

\maketitle

\begin{abstract}
	Instability and slowness are two main problems in deep reinforcement learning. Even if proximal policy optimization (PPO) is the state of the art, it still suffers from these two problems. We introduce an improved algorithm based on proximal policy optimization, mixed distributed proximal policy optimization (MDPPO), and show that it can accelerate and stabilize the training process. In our algorithm, multiple different policies train simultaneously and each of them controls several identical agents that interact with environments. Actions are sampled by each policy separately as usual, but the trajectories for the training process are collected from all agents, instead of only one policy. We find that if we choose some auxiliary trajectories elaborately to train policies, the algorithm will be more stable and quicker to converge especially in the environments with sparse rewards.
\end{abstract}


\section{Introduction}
Function approximation is now becoming a standard method to represent any policy or value function instead of tabular form in reinforcement learning \cite{sutton_reinforcement_2018}. An emerging trend of function approximation is to combine deep learning with much more complicated neural networks such as convolutional neural network (DQN \cite{mnih_playing_nodate}) in large-scale or continuous state. However, while DQN solves problems with high-dimensional state spaces, it can only handle discrete action spaces. Policy-gradient-like \cite{sutton2000policy} or actor-critic style \cite{konda_actor-critic_2000} algorithms directly model and train the policy. The output of a policy network is usually a continuous Gaussian distribution. However, compared with traditional lookup tables or linear function approximation, nonlinear function approximations cannot be guaranteed to convergence \cite{baird_residual_1995} due to its nonconvex optimization. Thus, the performance of algorithms using nonlinear value function or policy is very sensitive to the initial values of network weights.

Although PPO which we will mainly discuss in this paper is the state of the art, it is under the actor-critic framework, which has a very tricky disadvantage of instability during the training process. Combining with nonlinear function approximations makes the training even worse. Thus, the speed to convergence varies greatly or divergence may happen inevitably even if two policies have the same hyperparameters and environments, which makes developers confused about whether the reason is algorithm itself, hyperparameters or just initial values of neural network weights. So, for now, instability and slowness to convergence are still two significant problems in deep reinforcement learning with nonlinear function approximation \cite{bhatnagar_convergent_2009} and actor-critic style architecture.

In this paper, we propose a mixed distributed training approach which is more stable and quick to convergence, compared with standard PPO. We set multiple policies with different initial weights to control several agents at the same time. Decisions that affect the environments can only be made by each agent, but all trajectories are gathered and allocated to policies to train after elaborate selection. We define two selection criteria, (i) \emph{complete trajectories}, where an agent successfully completes a task like reaching a goal or lasting till the end of an episode, and (ii) \emph{auxiliary trajectories}, where an agent doesn't complete a task but the cumulative reward is much higher than other trajectories. In previous works, \cite{schulman_proximal_2017},\cite{nair_massively_2015},\cite{horgan_distributed_2018},\cite{espeholt2018impala} only distribute agents, not policies. \cite{mnih_asynchronous_2016} or RL with evolution strategies like \cite{salimans2017evolution} distribute policies, but they only choose and dispatch the best policy with the highest cumulative rewards. So, this work is the first, to our knowledge, to distribute both policies and agents and mix trajectories to stabilize and accelerate RL training.

For the purpose of simple implementation, we evaluated our approach on the Unity platform with Unity ML-Agents Toolkit \cite{juliani_unity:_2018} which is able to easily tackle with parallel physical scenes. Our algorithm performs very well in a Unity scenario, \emph{simple roller}, with random initial weights. The source code can be accessed at \url{https://github.com/BlueFisher/RL-PPO-with-Unity}.

\section{Related Work}\label{sec:related_work}
Several works have been done to solve the problem of instability. In deep Q network (DQN \cite{mnih_human-level_2015}), a separated neural network is used to generate the target Q value in online Q-learning update, making the training process much more stable. In order to break the correlation between transitions since the input of neural networks is preferably independent and identically distributed, the experience replay technique stores agent's transitions at each time step in a data set and samples a mini-batch from it during the training process. Deep deterministic policy gradient (DDPG \cite{lillicrap_continuous_2015}) applies both experience replay and target network into the deterministic policy gradient \cite{silver_deterministic_nodate}, and uses two different target networks to represent deterministic policy and Q value. However, in the aspect of stochastic policy gradient, almost all algorithms that based on actor-critic style \cite{konda_actor-critic_2000} have the problem of divergence, because the hardness of convergence of both actor and critic makes the whole algorithm even more unstable. Trust region policy optimization (TRPO \cite{schulman_trust_2015}) and proximal policy optimization (PPO \cite{schulman_proximal_2017},\cite{heess_emergence_2017}) try to bound the update of parameters between the policies before and after update, in order to make the extent of optimization not too large to out of control.

In terms of accelerating training process, the general reinforcement learning architecture (Gorila \cite{nair_massively_2015}) , asynchronous advantage actor critic (A3C \cite{mnih_asynchronous_2016}) and distributed PPO (DPPO \cite{heess_emergence_2017}) take the full use of multi-core processors and distributed systems. Multiple agents act in their own copy of the environments simultaneously. The gradients are computed separately and sent to a central parameter brain, which updates a central copy of the model asynchronously. The updated parameters will be then added and sent to every agent at a specific frequency. Distributed prioritized experience replay \cite{horgan_distributed_2018} overcomes the disadvantage that reinforcement learning cannot efficiently utilize the computing resource of GPU, but it only can be applied to the algorithm that uses replay buffer like DQN or DDPG. Besides, a distributed system is not easy and economical to implement. Reinforcement learning with unsupervised auxiliary tasks (UNREAL \cite{jaderberg_reinforcement_2016}) accelerates training by adding some additional tasks to learn. However, these tasks mainly focus on the tasks with graphical state input, which predict the change of pixels. And it is difficult to be transferred to some simple problems with vector state space environments.

\section{Background}\label{sec:background}
In reinforcement learning, algorithms based on policy gradient provide an outstanding paradigm for continuous action space problems. The purpose of all such algorithms is to maximize the cumulative expected rewards $L = \E\left[\sum_t r(s_t,a_t)\right]$. The most commonly used gradient of objective function $L$ with baseline can be written into the following form
\begin{equation}
	\begin{aligned}
		\nt L & = \int_\Sc \mu(s) \int_\Ac \nt\pt(a|s) A(s,a) \\
		      & = \E\left[\nt\log\pt(a|s) A(s,a)\right]
	\end{aligned}
\end{equation}
where $\Sc$ denotes the set of all states and $\Ac$ denotes the set of all actions respectively. $\mu(s)$ is an on-policy distribution over states. $\pi$ is a stochastic policy that maps state $s\in \Sc$ to action $a\in \Ac$, and $A$ is an advantage function.

TRPO replaces objective function $L$ with a surrogate objective, which has a constraint on the extent of update from an old policy to a new one. Specifically,
\begin{equation}
	\begin{aligned}
		\underset{\theta}{\mathrm{maximize}} \quad & \E\left[ \frac{\pt(a|s)}{\ptold(a|s)} A(s,a) \right]                   \\
		\mathrm{subject\ to} \quad                 & \E\big[ \mathrm{KL}[\ptold(\cdot | s),\pt(\cdot | s)] \big] \le \delta
	\end{aligned}
\end{equation}
where $\ptold$ is an old policy that generates actions but the parameters are fixed during the training process. $\pt$ is the policy that we try to optimize but not too much, so an upper bound $\delta$ is added to constrain the KL divergence between $\pt$ and $\ptold$.

PPO can be treated as an approximated but much simpler version of TRPO. It roughly clip the ratio between $\ptold$ and $\pt$ and change the surrogate objective function into the form
\begin{equation}\label{eq:clip}y
	L^{CLIP}_\theta = \E\left[ \min\Big( r(\theta)A, \mathrm{clip}\big( r(\theta),1-\epsilon,1+\epsilon \big)A \Big) \right]
\end{equation}
where $r(\theta) = \frac{\pt(a|s)}{\ptold(a|s)}$ and $\epsilon$ is the clipping bound. Note that PPO here is an actor-critic style algorithm, where actor is policy $\pt$ and critic is advantage function $A$.

In intuition, advantage function represents how good a state-action pair is compared with the average value of current state, i.e., $A(s,a)=Q(s,a)-V(s)$. The most used technique for computing advantage function is generalized advantage estimation (GAE \cite{schulman_high-dimensional_2016}). One popular style that can be easily applied to PPO or any policy-gradient-like algorithm is
\begin{equation}\label{eq:advantage}
	\begin{aligned}
		A_t(s_t,a_t) = r_t+\gamma r_{t+1} & + \cdots+\gamma^{T-t+1}r_{T-1} \\
		                                  & + \gamma^{T-t}V(s_T) - V(s_t)
	\end{aligned}
\end{equation}
where $T$ denotes the maximum length of a trajectory but not the terminal time step of a complete task, and $\gamma$ is a discounted factor. If the episode terminates, we only need to set $ V(s_T)$ to zero, without bootstrapping, which becomes $A_t = G_t - V(s_t)$ where $G_t$ is the discounted return following time $t$ defined in \cite{sutton_reinforcement_2018}. An alternative option is to use a more generalized version of GAE:
\begin{equation}\label{eq:gae}
	\begin{aligned}
		A_t(s_t,a_t) = \delta_t + (\gamma\lambda)\delta_{t+1} & + \cdots                              \\
		                                                      & + (\gamma\lambda)^{T-t+1}\delta_{T-1}
	\end{aligned}
\end{equation}
where $\delta_t=r_t+\gamma V(s_{t+1})-V(s_t)$ which is known as td-error. Noted that \eqref{eq:gae} will be reduced to \eqref{eq:advantage} when $\lambda=1$ but has high variance, or be reduced to $\delta_t$ when $\lambda=0$ but introduce bias.

We only need to approximate $\hat{V}_{\theta^v}(s)$ instead of $\hat{Q}(s,a)$ to compute the approximation of advantage function where $\hat{\delta}_t=r_t+\gamma \hat{V}_\tv(s_{t+1})-\hat{V}_\tv(s_t)$. So, instead of optimizing the surrogate objective, we have to minimize loss function
\begin{equation}
	\begin{aligned}
		L^V_\tv = \big( r_t+\gamma r_{t+1} & +\cdots+\gamma^{T-t+1}r_{T-1}                                  \\
		                                   & +\gamma^{T-t} \hat{V}_{\tvold}(s_T)-\hat{V}_{\tv}(s_t) \big)^2
	\end{aligned}
\end{equation}
where $\hat{V}_{\tvold}$ is the fixed value function which is used to compute $\hat{\delta}_t$.

\section{Mixed Distributed Proximal Policy Optimization}\label{sec:mdppo}
\begin{figure}
	\centering
	\includegraphics{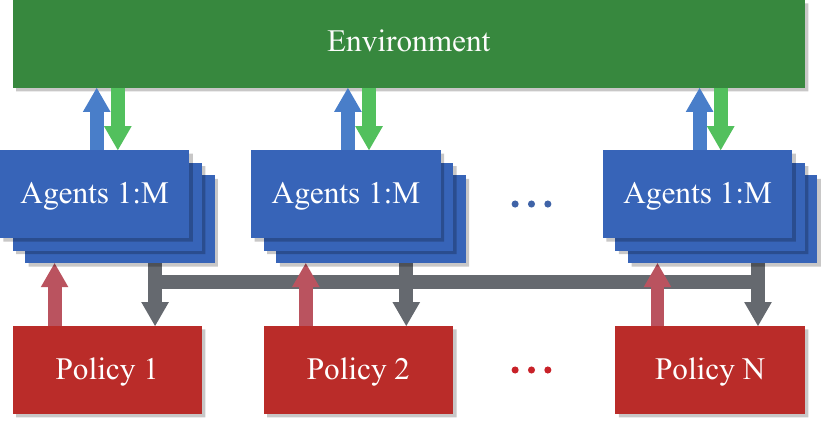}
	\caption{MDPPO: There are $N$ policies in the figure. Each policy controls $M$ agents (red lines), gives them the decision that how to interact with environments (blue and green lines). The data for computing gradients flows through gray lines where each policy gets updated not only by the data generated from agents that it controls, but also from other groups of agents.}
	\label{fig:mdppo}
\end{figure}

We distribute both policies and agents in contrast to DPPO where only agents are distributed. We set $N$ policies $\pi_1,\cdots ,\pi_N$ that have exactly the same architecture of policy network and value function network as well as hyperparameters. Each policy, like a command center, controls a group of $M$ identical agents $e^i_1,\cdots,e^i_M$ where $i$ denotes the serial number of policy that the group related to. So, $N\times M$ environments are running paralleled, interacting with $N\times M$ agents divided into $N$ groups. Each agent $e^i_j$ interacts its own shared policy $\pi_i$, sending states and receiving actions. When all agents have finished a complete episode $s_0, a_0, r_0, s_1 \cdots, s_T, a_T, r_t, s_{T+1}$, each policy is not only going to be fed with trajectories from the group of agents that it controls but also a bunch of ones from other groups elaborately that may accelerate and stabilize the training process. \figureref{fig:mdppo} illustrates how decisions and training data flows.

We define two sorts of trajectories being used to mixed training.

\textbf{Complete trajectories}, where an agent successfully completes a task. We simply divide tasks into two categories, (i) task with a specific goal, like letting an unmanned vehicle reach a designated target point, or making a robotic arm grab an object, and (ii) without a goal but trying to make the trajectory as long as possible till the maximum episode time is reached, like keeping a balance ball on a wobbling plate.

\textbf{Auxiliary trajectories}, where an agent does not complete a task but the cumulative reward is much higher than other trajectories. In many real-world scenarios, extrinsic rewards are extremely sparse. An unmanned surface vehicle may get no reward or cannot reach the target point for a long time in the beginning. However, we can utilize trajectories that have the highest cumulative rewards to encourage all policies to learn from current better performance.

\begin{algorithm}
	\caption{Mixed Distributed Proximal Policy Optimization (MDPPO)}
	\label{algo:mdppo}
	\For{$\mathrm{iteration}=1,2,\cdots$}{
	\For{$\mathrm{policy\ } \pi_i=\pi_1,\cdots ,\pi_N$}{
		\For{$\mathrm{environment\ }e^i_j=e^i_1,\cdots,e^i_M$}{
			Run policy $\pi_{\theta^i_\mathrm{old}}$ in environment $e^i_j$ for $T$ timesteps and compute advantage function. \\
			Store trajectories and cumulative rewards in dataset $D_i$.
		}
	}
	Compute \emph{complete} and \emph{auxiliary} trajectories from $D_1, \cdots ,D_N$ and store them in $D^-_1, \cdots ,D^-_N$. \\
	\For{$\mathrm{policy\ } \pi_i=\pi_1,\cdots ,\pi_N$}{
	optimize $L^{CLIP}_i$ (or $L^{CLIP^+}_i$) and $L^V_i$ respectively or combined $L^{CLIP+V}_i$ (or $L^{CLIP^+ +V}_i$) with mixed transitions from $D_i$ and $D^-_1, \cdots , D^-_{i-1}, D^-_{i+1}, \cdots , D^-_N$.
	}
	}
\end{algorithm}

We mix \emph{complete trajectories} and \emph{auxiliary trajectories}, and allocate them to every policy during the training process. So each policy computes gradient with its usual and mixed data altogether to update policy networks and value functions independently.

Note that in practice, mixed transitions from other policies may cause the denominator of ratio $r(\theta)$ in \eqref{eq:clip} to be zero if $\ptold(a|s)$ is too small that exceeds the floating range of computer. So that gradients cannot be computed correctly and result in "nan" problem when using deep learning frameworks like TensorFlow. We introduce four solutions, (i) simply removing all transitions that make $\ptold(a|s) < \varepsilon$ before training, (ii) adding a small number to denominator such that $r(\theta)=\pt(a|s)/(\ptold(a|s) + \epsilon)$ to make sure it has a lower bound, (iii) rewriting the ratio into exponential form $\exp(\pt(a|s) - \ptold(a|s))$ without modifying the clipping bound, and (iv), which we used in our experiments, rewriting the ratio into subtraction form instead of fractional one and limit the update of new policy to the range of $[-\epsilon,\epsilon]$:
\begin{equation}\label{eq:clip-}
	\begin{aligned}
		L^{CLIP^+}_\theta    & = \E\left[ \min\Big( r(\theta)A, \mathrm{clip}\big( r(\theta),-\epsilon,\epsilon \big)A \Big) \right], \\
		\mathrm{where \quad} & r(\theta) = \pt(a|s) - \ptold(a|s)
	\end{aligned}
\end{equation}
We no longer need to worry about "nan" problem since there is no denominator in \eqref{eq:clip-}.

Furthermore, there are two common forms of loss functions and network architectures. (i) training separated $L^V$ and $L^{CLIP}$ as the standard actor-critic style algorithms with completely different weights between policy and value function networks, and (ii) policy and value function networks can share a part of the same weights \cite{schulman_proximal_2017}. In this case, $L^V$ and $L^{CLIP}$ should be combined into one loss function, $L^{CLIP+V}=L^{CLIP} - L^V$, to make sure the update of shared weights is stable.

Note that the entropy term $L^S$ in \cite{schulman_proximal_2017} only acts as a regularizer, instead of maximizing the entropy \cite{Haarnoja2018Soft}. We discover that in our approach, the entropy term is optimal. Because a part of data used to update policy is actually generated by other different policies, which is the reason that MDPPO can encourage exploration. Although all policies' architectures and hyperparameters are identical, the initial weights of neural networks are totally different, which is the source of instability in standard PPO that gives MDPPO a more powerful ability of exploration. \algorithmref{algo:mdppo} illustrates the mixed distributed proximal policy optimization (MDPPO).

In contrast to DPPO where the computation of gradients is distributed to every agent and gradients are summarized to the center policy update then, in our approach, to take the full advantage of GPU resources, agents are only used to generate data, and transition data are transferred to the policy that it controls. So, gradients are computed in each policy centralized. In order to break the correlation between transitions, we tend to shuffle all transitions before training.

\subsection{MDPPO with Separated Critic}

\begin{figure}
	\centering
	\includegraphics[scale=1]{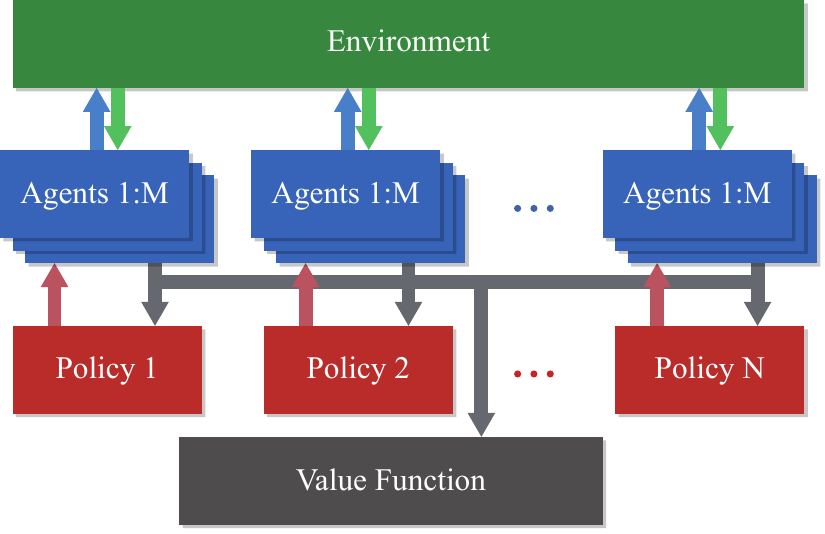}
	\caption{MDPPO with Separated Critic (MDPPOSC): The architecture is the same as Figure \ref{fig:mdppo}, except for the separated value function. The data through gray lines are now not only fed to actual $N$ policies, but to centralized value function (gray box), which needs all transitions generated by $N \times M$ agents without filtering.}
	\label{fig:mdppo_sep_critic}
\end{figure}

Actually, although the label of red box in \figureref{fig:mdppo} is shown as \emph{policy}, the data through gray lines should be used to update both policy and value function, which we usually call them actor and critic in actor-critic style algorithms. We find that having $N$ different value function networks is not quite necessary. We expect more actors and mix them to some extent to encourage exploration and increase diversity, which may speed up the training process and make it more stable. However, critic only needs state-reward transitions as much as possible, trying to make the approximation of states' value more correct. Besides, the loss function $L^V_t = (G(s_t) - \hat{V}(s_t))^2$ is irrelevant to the policy or action. So, we can separate  the value function network from \figureref{fig:mdppo} and feed all transitions to one centralized critic. \figureref{fig:mdppo_sep_critic} shows the separated value function which is extracted from the policies in \figureref{fig:mdppo}.

However, under the circumstances of sparse positive rewards, the value function may be fed with a large number of useless transitions generated from a bad policy at the beginning, which will cause the critic to converge to a local optimum in a very short time and hard to jump out the gap, since the algorithm would have learned a terrible policy with the guide of a terrible advantage function. In practice, we only simply feed transitions where td-error is greater than a threshold. We also set the threshold to exponential decay since we would like value function to explore more possibilities at the beginning, but converge to the real optimal point at the end.

\section{Experiments}\label{sec:experiments}

Distributed experiment environments are difficult and expensive to build in the physical world. Even in one computer, it is still not easy to implement a parallel system which has to manage multiple agents, threads and physical simulations. However, as a game engine, Unity can provide us with fast and distributed simulation. What we only need to do is copying an environment several times, Unity will run all environments in parallel without any extra coding.

So we test our approach in a standard scenario, \emph{Simple Roller}, on the Unity platform with Unity ML-Agents Toolkit \cite{juliani_unity:_2018}. The agent's goal is to hit the target as quickly as possible while avoiding falling down the floor plane.

\begin{figure*}
	\centering
	\begin{subfigure}{0.31\textwidth}
		\centering
		\includegraphics[width=\textwidth]{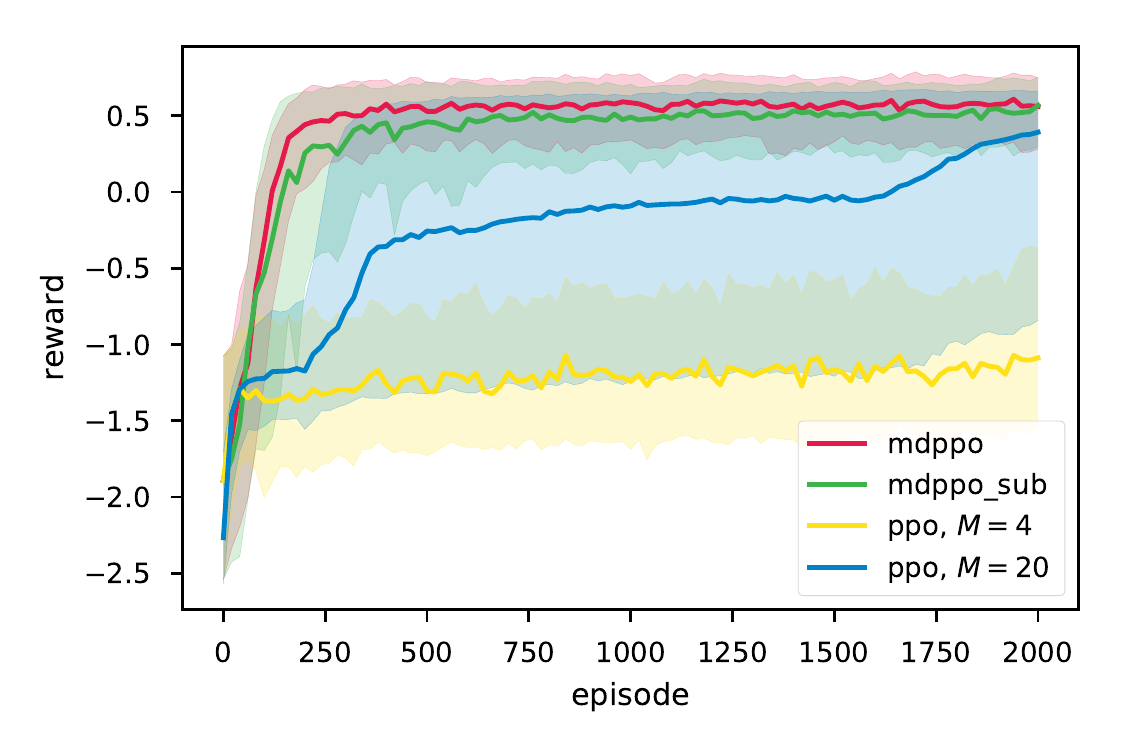}
		\caption{MDPPO (separated)}
		\label{fig:simple_roller_sep}
	\end{subfigure}
	\begin{subfigure}{0.31\textwidth}
		\centering
		\includegraphics[width=\textwidth]{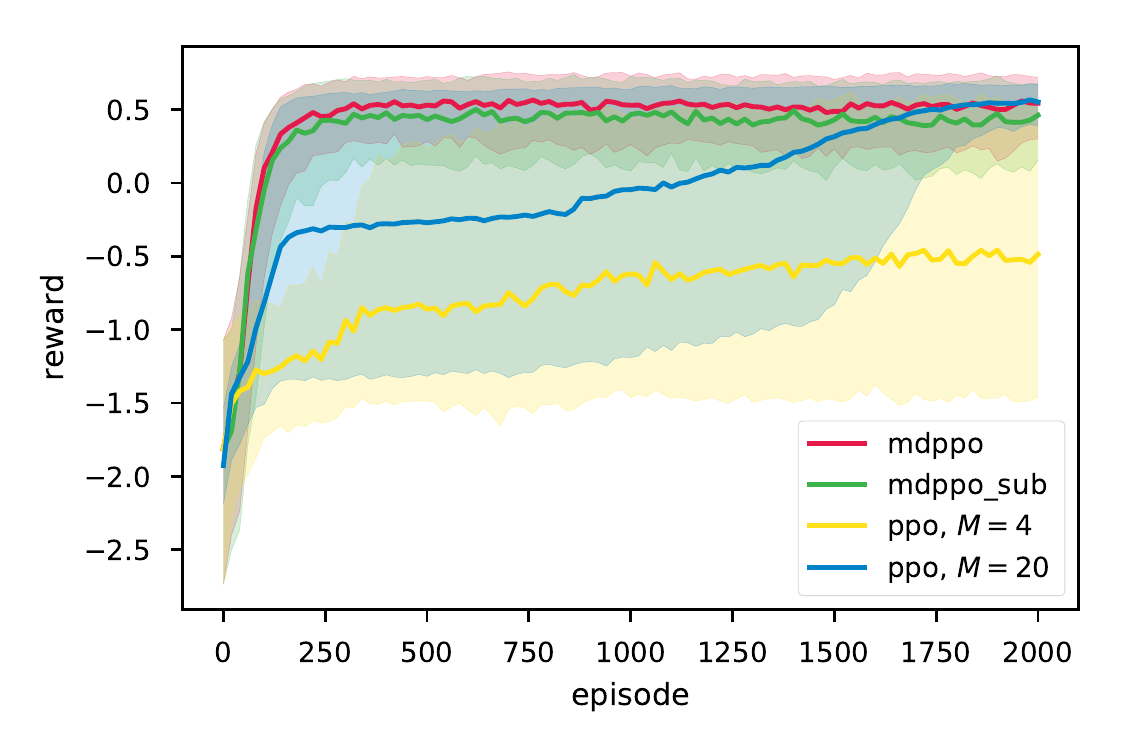}
		\caption{MDPPO (shared)}
		\label{fig:simple_roller_std}
	\end{subfigure}
	\begin{subfigure}{0.31\textwidth}
		\centering
		\includegraphics[width=\textwidth]{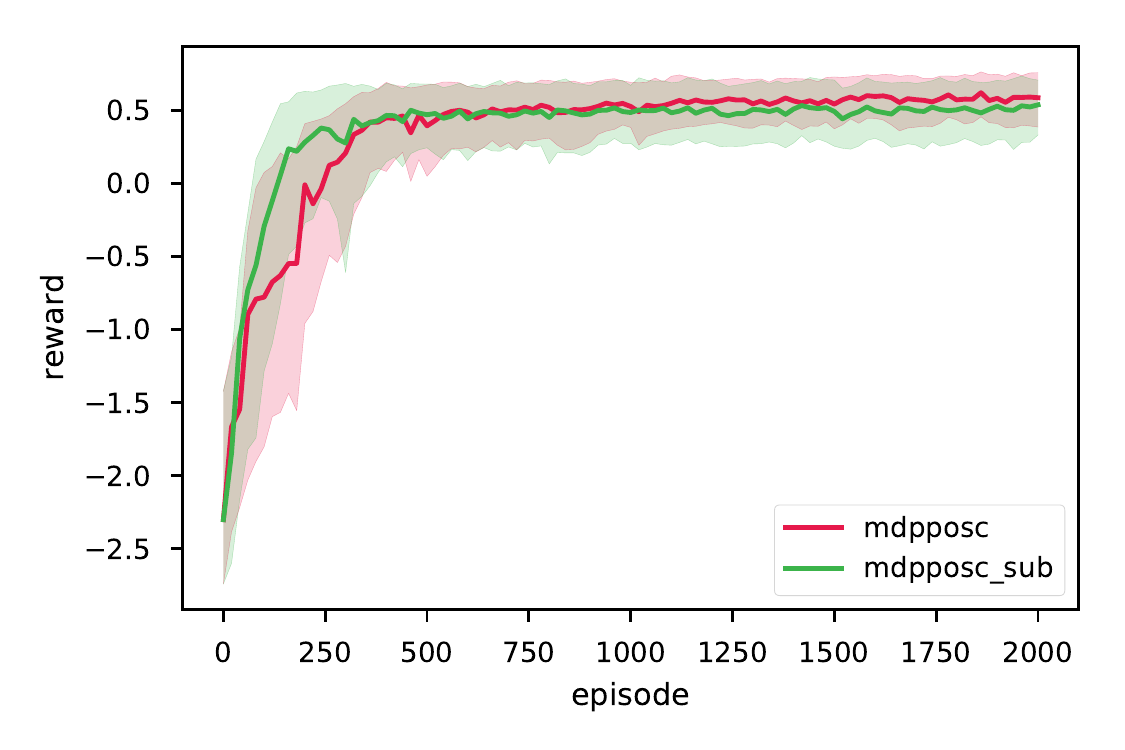}
		\caption{MDPPOSC}
		\label{fig:simple_roller_sep_critic}
	\end{subfigure}

	\caption{In \ref{fig:simple_roller_sep} and \ref{fig:simple_roller_std}, MDPPO (red) and MDPPO with subtraction ratio (green) are tested with 5 policies. Each policy has 4 agents. Yellow and blue lines are standard PPO algorithms with 4 agents and 20 agents respectively. PPO with 4 agents are tested 5 times, meanwhile PPO with 20 agents are tested 10 times. The lighter color in the background shows the best and worst performance of each algorithm test. The more narrow the background, the more stable the algorithm. \ref{fig:simple_roller_sep_critic} shows MDPPO with separated critic. The performance of two algorithms is similar and outperform standard PPO both.}
\end{figure*}

The first experiment is under 5 policies, i.e., $N=5$. Each policy is modeled as a multivariate Gaussian distribution with a mean vector and a diagonal covariance matrix, parameterized by $\theta=\{\theta_\mu\in\mathbb{R}^2, \theta_\Sigma\in\mathbb{R}^2\}$. We set 20 distributed agents and environments in all, i.e., each policy controlls 4 agents and $M=4$. The \emph{complete} trajectories are simply regarded as the transitions that rollers hit targets. As for \emph{auxiliary} trajectories, we choose the top 40\% transitions that have the greatest cumulative rewards. \figureref{fig:simple_roller_sep} and \figureref{fig:simple_roller_std} shows our four experiments, MDPPO, MDPPO with subtraction ratio, standard PPO with four distributed agents and twenty distributed agents. We divide them into two groups, one with shared weights and the other not.

Both MDPPO and MDPPO with subtraction ratio outperform standard PPO. The time to convergence is shorter, and most importantly, MDPPO is much more stable. All five policies are converged almost simultaneously and have similar performance. In contrast, only a few policies that optimized by standard PPO have converged, which shows the instability of PPO. Besides, the speed to convergence of some policies is quite different. Some converge as quickly as MDPPO, but most of them delay much, even not converge. We cannot find a pattern of time to convergence with different initial parameters for now.

We also test the performance of our approach with different combinations of $N$ and $M$, but the total number of agents is set to 20, i.e., $N\times M = 20$. We find that with the decrease of $N$, the algorithm will be more unstable, especially when $N=1$, i.e., standard PPO. Besides, even the algorithm that has the best performance is slower to convergence than any other algorithms that $N>1$. However, it is also not always good to increase $N$. In this experiment, a larger $N$ may indeed accelerate the training process, but it will also make the algorithm much more unstable.

Finally, we test MDPPO with separated critic. Figure \ref{fig:simple_roller_sep_critic} illustrates the results of algorithms with two ratio forms. Both show similar great performance.

\section{Conclusion}\label{sec:conclusion}

In this paper, we present two algorithms. Mixed Distributed Proximal Policy Optimization (MDPPO) and Mixed Distributed Proximal Policy Optimization with Separated Critic (MDPPOSC) and make three contributions, (i), a method that distributes not only agents but also policies. Multiple policies train simultaneously and each of them controlls multiple identical agents. (ii), the transitions are mixed from all policies and fed to each policy for training after elaborate selection. We define \emph{complete trajectories} and \emph{auxiliary trajectories} to accelerate the training process, (iii), we formulate a practical MDPPO algorithm based on the Unity platform with Unity ML-Agents Toolkit to simplify the implementation of distributed reinforcement learning system. Our experiments indicate that MDPPO is much more robust under any random initial neural network weights and could accelerate the training process compared with standard PPO.

\section*{Acknowledgment}

The research reported in this paper is supported in part by the National Natural Science Foundation of China (No. 91746203, No. 61802245 and No. U1813217), and the project of the Intelligent Ship Situation Awareness System.

\bibliographystyle{plain}
\bibliography{mdppo}

\begin{thebibliography}{10}

\bibitem{baird_residual_1995}
Leemon Baird.
\newblock Residual algorithms: Reinforcement learning with function
  approximation.
\newblock In {\em Machine Learning Proceedings 1995}, pages 30--37. Elsevier.

\bibitem{bhatnagar_convergent_2009}
Shalabh Bhatnagar, Doina Precup, David Silver, Richard~S Sutton, Hamid~R. Maei,
  and Csaba Szepesvári.
\newblock Convergent temporal-difference learning with arbitrary smooth
  function approximation.
\newblock In Y.~Bengio, D.~Schuurmans, J.~D. Lafferty, C.~K.~I. Williams, and
  A.~Culotta, editors, {\em Advances in Neural Information Processing Systems
  22}, pages 1204--1212. Curran Associates, Inc.

\bibitem{espeholt2018impala}
Lasse Espeholt, Hubert Soyer, Remi Munos, Karen Simonyan, Volodymir Mnih, Tom
  Ward, Yotam Doron, Vlad Firoiu, Tim Harley, Iain Dunning, et~al.
\newblock Impala: Scalable distributed deep-rl with importance weighted
  actor-learner architectures.
\newblock {\em arXiv preprint arXiv:1802.01561}, 2018.

\bibitem{Haarnoja2018Soft}
Tuomas Haarnoja, Aurick Zhou, Pieter Abbeel, and Sergey Levine.
\newblock Soft actor-critic: Off-policy maximum entropy deep reinforcement
  learning with a stochastic actor.
\newblock 2018.

\bibitem{heess_emergence_2017}
Nicolas Heess, Dhruva {TB}, Srinivasan Sriram, Jay Lemmon, Josh Merel, Greg
  Wayne, Yuval Tassa, Tom Erez, Ziyu Wang, S.~M.~Ali Eslami, Martin Riedmiller,
  and David Silver.
\newblock Emergence of locomotion behaviours in rich environments.

\bibitem{horgan_distributed_2018}
Dan Horgan, John Quan, David Budden, Gabriel Barth-Maron, Matteo Hessel, Hado
  van Hasselt, and David Silver.
\newblock Distributed prioritized experience replay.

\bibitem{jaderberg_reinforcement_2016}
Max Jaderberg, Volodymyr Mnih, Wojciech~Marian Czarnecki, Tom Schaul, Joel~Z.
  Leibo, David Silver, and Koray Kavukcuoglu.
\newblock Reinforcement learning with unsupervised auxiliary tasks.

\bibitem{juliani_unity:_2018}
Arthur Juliani, Vincent-Pierre Berges, Esh Vckay, Yuan Gao, Hunter Henry,
  Marwan Mattar, and Danny Lange.
\newblock Unity: A general platform for intelligent agents.

\bibitem{konda_actor-critic_2000}
Vijay~R Konda and John~N Tsitsiklis.
\newblock Actor-critic algorithms.
\newblock In {\em Advances in neural information processing systems}, pages
  1008--1014.

\bibitem{lillicrap_continuous_2015}
Timothy~P. Lillicrap, Jonathan~J. Hunt, Alexander Pritzel, Nicolas Heess, Tom
  Erez, Yuval Tassa, David Silver, and Daan Wierstra.
\newblock Continuous control with deep reinforcement learning.

\bibitem{mnih_asynchronous_2016}
Volodymyr Mnih, Adrià~Puigdomènech Badia, Mehdi Mirza, Alex Graves,
  Timothy~P. Lillicrap, Tim Harley, David Silver, and Koray Kavukcuoglu.
\newblock Asynchronous methods for deep reinforcement learning.

\bibitem{mnih_playing_nodate}
Volodymyr Mnih, Koray Kavukcuoglu, David Silver, Alex Graves, Ioannis
  Antonoglou, Daan Wierstra, and Martin Riedmiller.
\newblock Playing atari with deep reinforcement learning.
\newblock page~9.

\bibitem{mnih_human-level_2015}
Volodymyr Mnih, Koray Kavukcuoglu, David Silver, Andrei~A. Rusu, Joel Veness,
  Marc~G. Bellemare, Alex Graves, Martin Riedmiller, Andreas~K. Fidjeland,
  Georg Ostrovski, Stig Petersen, Charles Beattie, Amir Sadik, Ioannis
  Antonoglou, Helen King, Dharshan Kumaran, Daan Wierstra, Shane Legg, and
  Demis Hassabis.
\newblock Human-level control through deep reinforcement learning.
\newblock 518(7540):529--533.

\bibitem{nair_massively_2015}
Arun Nair, Praveen Srinivasan, Sam Blackwell, Cagdas Alcicek, Rory Fearon,
  Alessandro De~Maria, Vedavyas Panneershelvam, Mustafa Suleyman, Charles
  Beattie, Stig Petersen, and {others}.
\newblock Massively parallel methods for deep reinforcement learning.

\bibitem{salimans2017evolution}
Tim Salimans, Jonathan Ho, Xi~Chen, Szymon Sidor, and Ilya Sutskever.
\newblock Evolution strategies as a scalable alternative to reinforcement
  learning.
\newblock {\em arXiv preprint arXiv:1703.03864}, 2017.

\bibitem{schulman_trust_2015}
John Schulman, Sergey Levine, Philipp Moritz, Michael~I. Jordan, and Pieter
  Abbeel.
\newblock Trust region policy optimization.

\bibitem{schulman_high-dimensional_2016}
John Schulman, Philipp Moritz, Sergey Levine, Michael~I Jordan, and Pieter
  Abbeel.
\newblock {HIGH}-{DIMENSIONAL} {CONTINUOUS} {CONTROL} {USING} {GENERALIZED}
  {ADVANTAGE} {ESTIMATION}.
\newblock page~14.

\bibitem{schulman_proximal_2017}
John Schulman, Filip Wolski, Prafulla Dhariwal, Alec Radford, and Oleg Klimov.
\newblock Proximal policy optimization algorithms.

\bibitem{silver_deterministic_nodate}
David Silver, Guy Lever, Nicolas Heess, Thomas Degris, Daan Wierstra, and
  Martin Riedmiller.
\newblock Deterministic policy gradient algorithms.
\newblock page~9.

\bibitem{sutton_reinforcement_2018}
Richard~S Sutton and Andrew~G Barto.
\newblock {\em Reinforcement learning: An introduction}.
\newblock {MIT} press.

\bibitem{sutton2000policy}
Richard~S Sutton, David~A McAllester, Satinder~P Singh, and Yishay Mansour.
\newblock Policy gradient methods for reinforcement learning with function
  approximation.
\newblock In {\em Advances in neural information processing systems}, pages
  1057--1063, 2000.

\end{thebibliography}

\end{document}